\documentclass[letterpaper, 10 pt, conference]{ieeeconf}  

\IEEEoverridecommandlockouts                              

\usepackage{graphics} 
\usepackage{epsfig} 
\usepackage{mathptmx} 
\usepackage{times} 
\usepackage{amsmath} 
\usepackage{amssymb}  
\usepackage{caption}
\usepackage{subcaption}
\usepackage[hidelinks]{hyperref}
\usepackage{cleveref}

\usepackage{graphicx}
\usepackage{epstopdf}
\usepackage{tikz-cd}

\usepackage{tcolorbox}

\usepackage{listings}
\usepackage{xcolor}  
\crefname{lstlisting}{snippet}{snippets}
\Crefname{lstlisting}{Snippet}{Snippets}

\definecolor{rustbg}{rgb}{0.15,0.15,0.15}          
\definecolor{rustkeyword}{rgb}{0.0,0.5,0.0}        
\definecolor{rusttype}{rgb}{0.8, 0.1, 0.5}     
\definecolor{rustcomment}{rgb}{0.458,0.443,0.368} 
\definecolor{ruststring}{rgb}{0.9,0.86,0.45}      
\definecolor{rustnumber}{rgb}{0.68,0.5,1.0}       
\definecolor{rusttext}{rgb}{0.97,0.97,0.95}       

\lstdefinelanguage{Rust}{
  keywords=[1]{fn, let, mut, if, else, match, while, loop, for, in, break, continue, return, struct, enum, mod, pub, impl, trait, const, static, use, crate, ref, as, move, async, await, dyn, where, Self, self, super, type},
  keywords=[2]{u8, u16, i32, i64, u32, u64, f32, f64, usize, isize, bool, char, str, String, Option, Result, Vec, Box},
  sensitive=true,
  comment=[l]{//},
  morecomment=[s]{/*}{*/},
  morestring=[b]",
  alsoletter={_:},
}

\lstdefinestyle{ruststyle}{
  language=Rust,
  commentstyle=\color{rustcomment}\itshape,
  keywordstyle=[1]\color{rustkeyword}\bfseries,
  keywordstyle=[2]\color{rusttype}\bfseries,
  stringstyle=\color{rusttype},
  basicstyle=\ttfamily\small,
  breaklines=true,
  breakatwhitespace=true,
  showstringspaces=false,
  captionpos=b,
}
\lstdefinestyle{monokai}{
  backgroundcolor=\color{black!90},   
  basicstyle=\footnotesize\ttfamily\color{white},
  keywordstyle=\color{orange},
  stringstyle=\color{green!70!black},
  commentstyle=\color{gray!60},
  numberstyle=\tiny\color{gray},
  breaklines=true,
  frame=none,
  showstringspaces=false,
  tabsize=2,
  morekeywords={sudo, apt, install, bash},
}

\newtcolorbox{installbox}{
  colback=blue!3!white,      
  colframe=blue!70!black,    
  coltitle=black,
  fontupper=\bfseries,
  boxrule=0.5mm,
  arc=3mm,                   
  top=2mm,
  bottom=2mm,
  left=3mm,
  right=3mm,
  before upper={\vspace{2pt}},
  after upper={\vspace{2pt}},
}

\title{\LARGE \bf
Kornia-rs: A Low-Level 3D Computer Vision Library In Rust
}

\author{
    Edgar Riba$^{1}$, Jian Shi$^{1,2}$, Aditya Kumar$^{3}$, Andrew Shen$^{3}$ and Gary Bradski$^{1,4}$%
    \thanks{$^{1}$Kornia AI Research Organization
    }%
    \thanks{$^{2}$King Abdullah University of Science and Technology}%
    \thanks{$^{3}$External contributors collaborating on this project}%
    \thanks{$^{4}$OpenCV.org}%
}

\begin{document}

\maketitle
\thispagestyle{empty}
\pagestyle{empty}

\begin{abstract}

We present \texttt{kornia-rs}, a high-performance 3D computer vision library written entirely in native Rust, designed for safety-critical and real-time applications. Unlike C++-based libraries like OpenCV or wrapper-based solutions like OpenCV-Rust, \texttt{kornia-rs} is built from the ground up to leverage Rust’s ownership model and type system for memory and thread safety. \texttt{kornia-rs} adopts a statically-typed tensor system and a modular set of crates, providing efficient image I/O, image processing and 3D operations. To aid cross-platform compatibility, \textit{kornia-rs} offers Python bindings, enabling seamless and efficient integration with Rust code. Empirical results show that \textit{kornia-rs} achieves a $3\sim5\times$ speedup in image transformation tasks over native Rust alternatives, while offering comparable performance to C++ wrapper-based libraries.  In addition to 2D vision capabilities, \textit{kornia-rs} addresses a significant gap in the Rust ecosystem by providing a set of 3D computer vision operators. This paper presents the architecture and performance characteristics of \textit{kornia-rs}, demonstrating its effectiveness in real-world computer vision applications.

\end{abstract}

\section{INTRODUCTION}

The Rust~\cite{matsakis2014rust} programming language has emerged as a powerful tool for system programming, offering unique guarantees of memory safety and thread safety through its ownership model and type system. While the Computer Vision and 3D landscape has traditionally been dominated by C++ and Python libraries such as OpenCV~\cite{opencv_library}, Pillow~\cite{Murray2025-dd}, Scikit Image~\cite{scikit-image}, Open3D~\cite{zhou2018open3dmodernlibrary3d}, Point Cloud Library (PCL)~\cite{Rusu_ICRA2011_PCL}, or Kornia~\cite{eriba2019kornia}, the Rust ecosystem has seen growing interest in developing high-performance, safe alternatives. However, existing Rust-based CV libraries such as \textit{OpenCV-Rust} and \textit{tch-rs} (PyTorch~\cite{paszke2019pytorch} bindings) are primarily wrappers around C++ libraries, which are challenging to deploy on different architectures such as ARM or WebAssembly for edge computing and web-based applications. Nevertheless, the Rust-native libraries such as \textit{image-rs}~\cite{image-rs} and \textit{ndarray}~\cite{ndarry-rs} provide limited functionality for image processing. There remains a need for a comprehensive computer vision library that fully leverages Rust's safety features while delivering competitive performance for modern computer vision tasks (\textit{e.g.} geometry, ML inference).

This paper presents \textit{kornia-rs}, a novel Open Source Computer Vision library under Apache-2 License that bridges this gap by combining Rust's safety guarantees with modern computer vision capabilities. Using Rust's ownership model, \textit{kornia-rs} ensures that image data and tensor operations are handled safely without runtime overhead. This makes \textit{kornia-rs} particularly suitable for applications where reliability and performance are critical, such as embedded systems, robotics, or real-time processing pipelines. Unlike traditional computer vision libraries that rely on runtime checks or garbage collection, \textit{kornia-rs} leverages Rust's compile-time guarantees to prevent common programming errors while maintaining high performance. The library's modular architecture enables seamless integration with other Rust-based frameworks, \textit{e.g.} \textit{copper-rs}~\cite{copper-rs} and \textit{dora-rs}~\cite{dora-rs} for robotics applications, and integrates with neural network inference engines such as \textit{candle}~\cite{candle} and \textit{ort}~\cite{ort}.


This paper first explores \textit{kornia-rs}'s design choices to achieve competitive performances against previous Rust-based computer vision libraries. Our results present to be $3\sim 5$ times faster for image transformations against Rust-native libraries, while maintaining comparative against those C++ wrapper-based libraries. Secondly, \textit{kornia-rs} fills the gap for the lacks of a 3D computer vision operators in rust community. Lastly, we present integration examples to further demonstrate the usability and accessibility of our library.
Users can easily access \textit{kornia-rs} through GitHub~\footnote{\url{https://github.com/kornia/kornia-rs.git}} or install \textit{kornia-rs} by:
\begin{installbox}
\lstinline[language=bash]{cargo install kornia}
\end{installbox}
\vspace{1em}


\section{Design principles}

Benefit from compelling foundation, offered by the Rust~\cite{matsakis2014rust} programming language, for building efficient, safe, and composable systems, \textit{kornia-rs} delivers a low-level library that addresses the performance and correctness requirements for computer vision applications with strong compile-time guarantees.


\subsection{Modular Architecture}
The library is organized as a collection of independent but interoperable crates, each encapsulating a specific functional domain of computer vision. This modular decomposition enables fine-grained dependency control, reduces compilation overhead, and facilitates integration into constrained embedded or robotic systems. The main but not limited crates include:

\begin{itemize}
    \item \texttt{kornia-tensor}: Provides generic, type-safe tensor abstractions with core tensor operations and memory management.
    \item \texttt{kornia-image}: Defines image-specific data structures and operations.
    \item \texttt{kornia-io}: Implements image and video I/O operations with support for multiple codecs.
    \item \texttt{kornia-imgproc}: Contains core image processing algorithms such as filtering and geometric transforms.
    \item \texttt{kornia-3d}: Supports 3D computer vision operations, including pointclouds I/O and 3D geometry.
    \item \texttt{kornia-icp}: Provides a standalone implementation of the Iterative Closest Point algorithm.
\end{itemize}

\subsection{Core API Design}

A key principle of the API is minimizing runtime error surfaces by taking the advantage of the Rust's type system. Errors due to invalid formats, Image and Tensor rank mismatches, or unsupported operations are surfaced at compile time through trait bounds and strongly typed interfaces. Optional features such as JPEG decoding or Gstreamer support are exposed via feature flags, further reducing runtime surprises and enabling minimal builds. The gist features are showing in~\Cref{listing:main}.

\begin{lstlisting}[language=rust, basicstyle=\footnotesize\ttfamily,
caption={An example that illustrates several core ideas of \textit{kornia-rs}: 1) \textbf{Compile-time tensor rank:} By encoding dimensions in the type system (e.g., \texttt{Tensor<T, N, \_>}), we catch rank mismatches at compile time, unlike Python's NumPy or PyTorch. 2) \textbf{Allocator abstraction:} \textit{CpuAllocator} is interchangeable with other backends (\textit{e.g.}, GPU allocators in future work) without changing tensor logic. 3) \textbf{No implicit allocations:} All ops require the user to manage memory explicitly. This removes hidden costs and helps in embedded or real-time systems.},
label={listing:main}]
use kornia::tensor::Tensor;
use kornia::tensor_ops::TensorOps;

// Create a tensor from a slice
let data: [f32; 4] = [1f32, 2, 3, 4];
let x = Tensor::<u8, 2, _>::from_shape_slice(
    [2, 2], &data, CpuAllocator)?;

// Create another tensor from single value
let y = Tensor::<f32, 2, _>::from_shape_val(
    [2, 2], 1.0, CpuAllocator);

// Calculate mean sqaure error (x - y)^2
let mse = x.sub(&y)?.powf(2.0).mean()?;
\end{lstlisting}

\subsubsection{\textbf{Typed Tensor Abstraction} (\texttt{kornia-tensor})}

At the core of the library lies a generic \lstinline[language=Rust]{Tensor<T, N>} type, where \texttt{T} denotes the element type (e.g., \texttt{f32}, \texttt{u8}) and \texttt{N} encodes the tensor's rank as a compile-time constant. This design makes tensor shapes explicit and verifiable at compile time. For instance, a 3D image tensor is represented as \texttt{Tensor<f32, 3>}, reducing ambiguity and runtime shape mismatches.

\subsubsection{\textbf{Image as a Strongly-Typed Struct} (\texttt{kornia-image})}

Building on top of the tensor abstraction, the \texttt{Image<T, C>} type introduces a foundational Image API that can define different image type based on the bit depth and the number of channels, e.g. from grayscale or color images to depth maps. This enables compile-time safety for image operations, such as ensuring color conversions or filters are applied only to valid formats, and allows specialization for performance optimizations.
This approach contrasts against previous Rust image processing libraries (e.g., \textit{image-rs}, \textit{ndarray}), which typically rely on runtime assertions.

\begin{lstlisting}[language=rust, basicstyle=\footnotesize\ttfamily,caption={User can define their own image types based on their needs and applications.},label={listing:image}]
use kornia::image::Image;

type ImageRgb8 = Image<u8, 3>;
type ImageGray8 = Image<u8, 1>;
type ImageRgbd = Image<u16, 4>;
\end{lstlisting}

\subsubsection{\textbf{Zero-Copy Views and Ownership Semantics}}

Leveraging Rust's ownership and borrowing system, \textit{kornia-rs} avoids unnecessary memory allocations through zero-copy view types. Operations like cropping or slicing return lightweight, borrow-checked views rather than duplicating data. This approach maintains memory safety while enabling real-time performance, a significant advantage over garbage-collected or reference-counted environments.

\subsubsection{\textbf{Pre-Allocated Buffers for Real-Time Execution}}

To meet the demands of real-time systems, \textit{kornia-rs} provides APIs that operate on pre-allocated memory buffers. This design allows developers to avoid heap allocations during performance-critical execution loops, enabling predictable latency and high throughput—key requirements in robotics, embedded vision, and autonomous systems. A larger chunk of the functionality found in the library take user-provided output buffers as explicit arguments, ensuring memory reuse and eliminating hidden allocations. Combined with Rust’s ownership model, these patterns enable deterministic memory management without sacrificing safety.
Moreover, users can construct memory arenas or scratch buffers once and pass mutable references throughout the pipeline.






\begin{figure*}[t!]
    \centering
    \begin{subfigure}[t]{.99\textwidth}
        \centering
        \includegraphics[width=.24\textwidth]{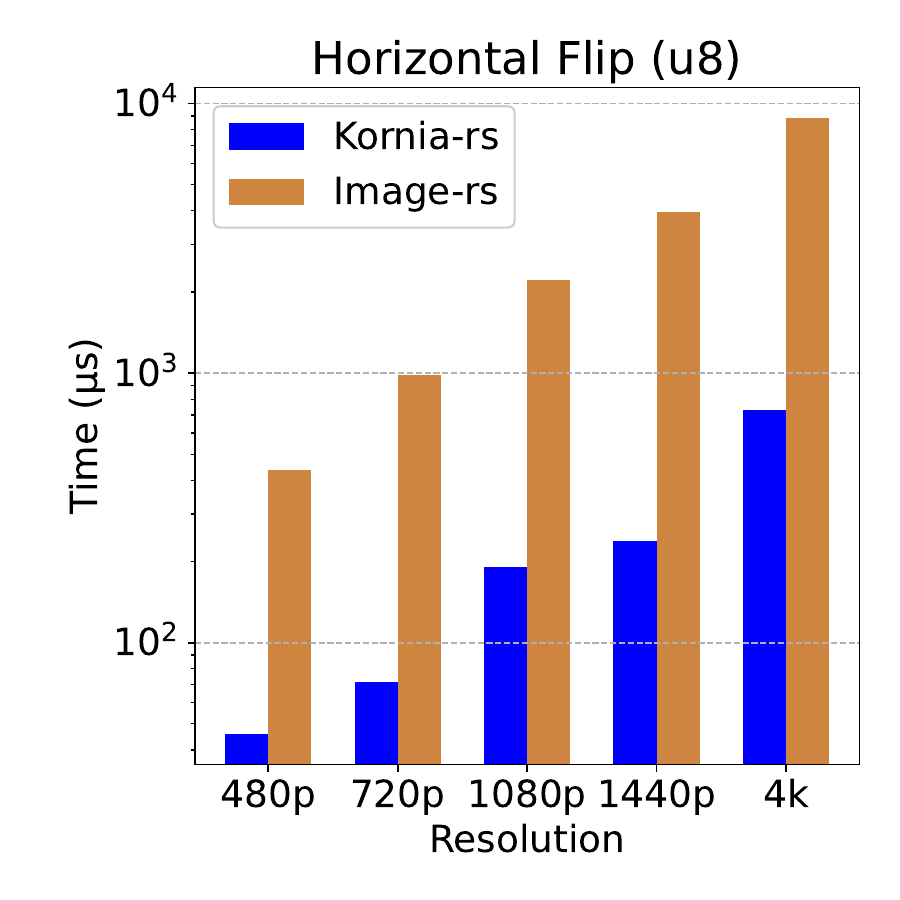}
        \includegraphics[width=.24\textwidth]{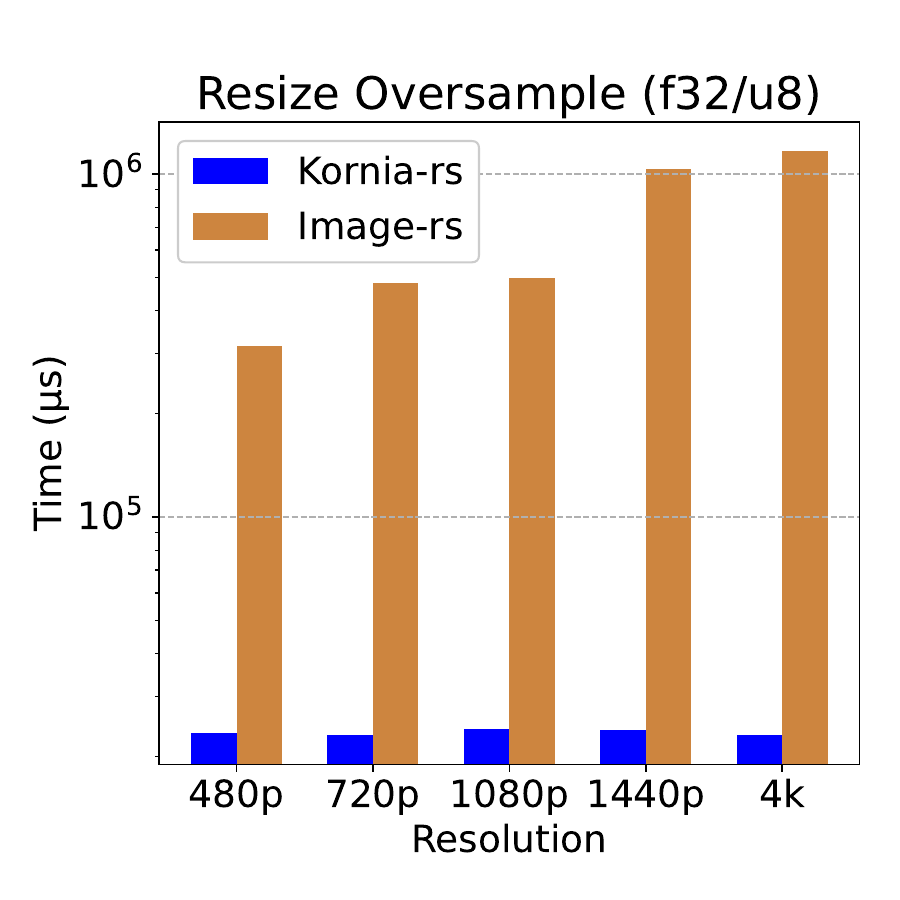}
        \includegraphics[width=.24\textwidth]{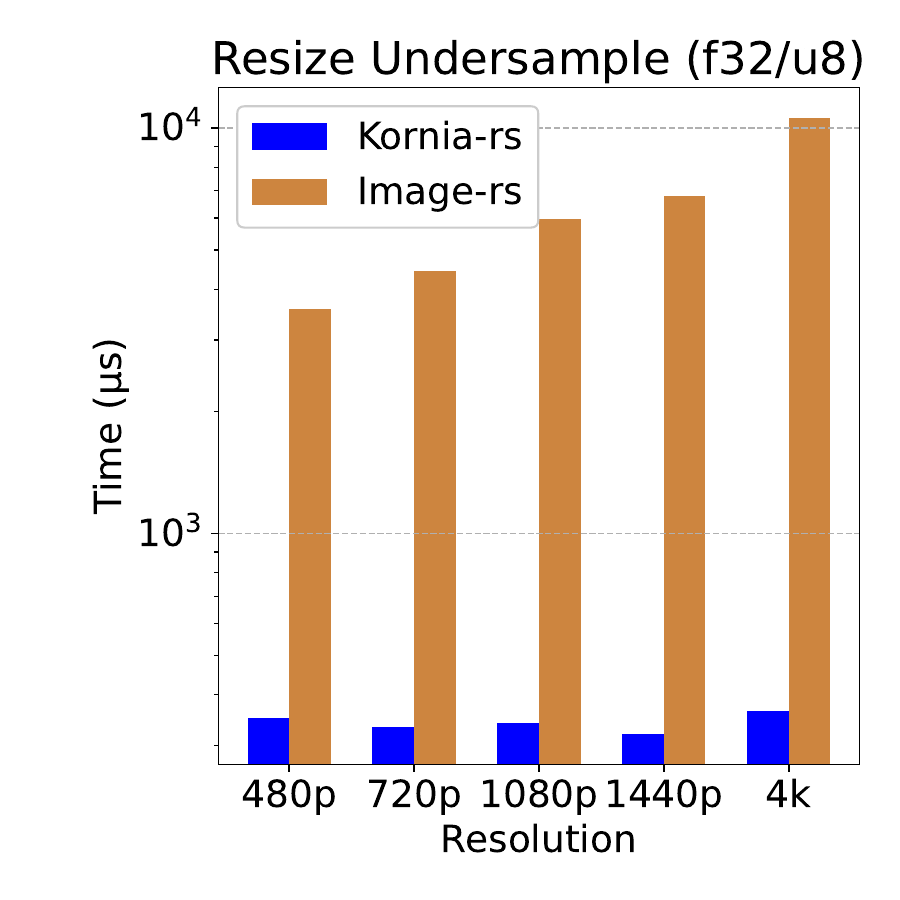}
        \includegraphics[width=.24\textwidth]{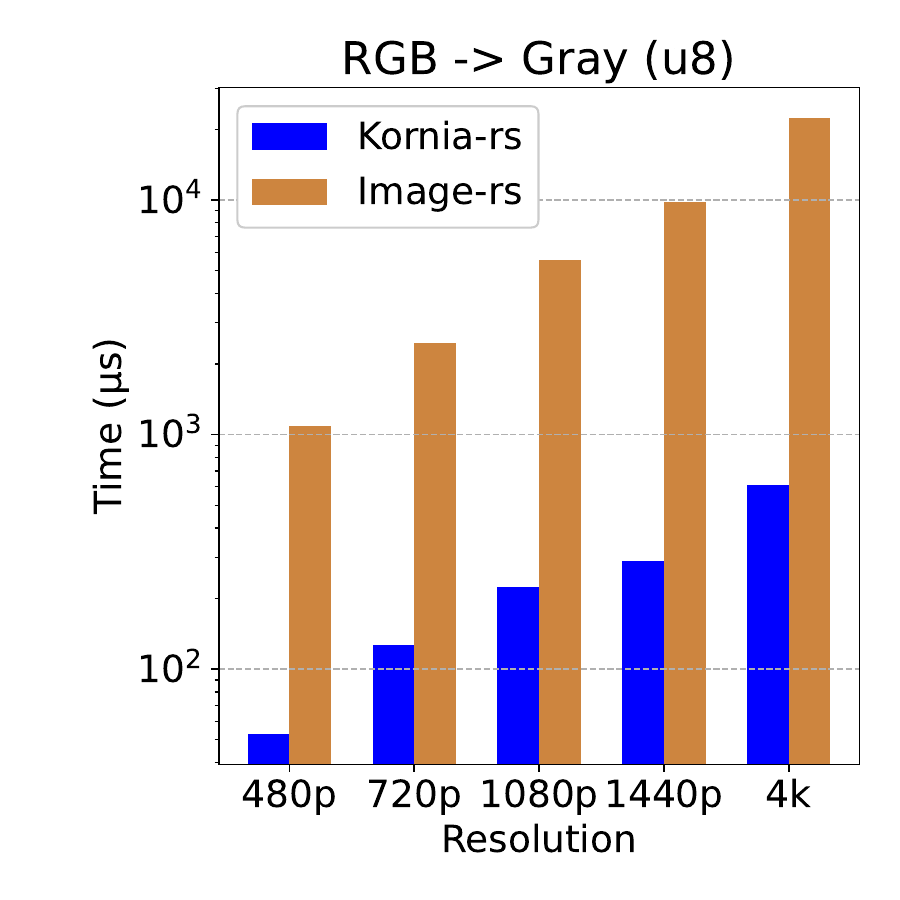}
        \caption{Comparison using the Rust-native functionality}
        \label{fig:rust-benchmarks}
    \end{subfigure}%
    \\
    \begin{subfigure}[t]{.99\textwidth}
        \centering
        \includegraphics[width=.3\textwidth]{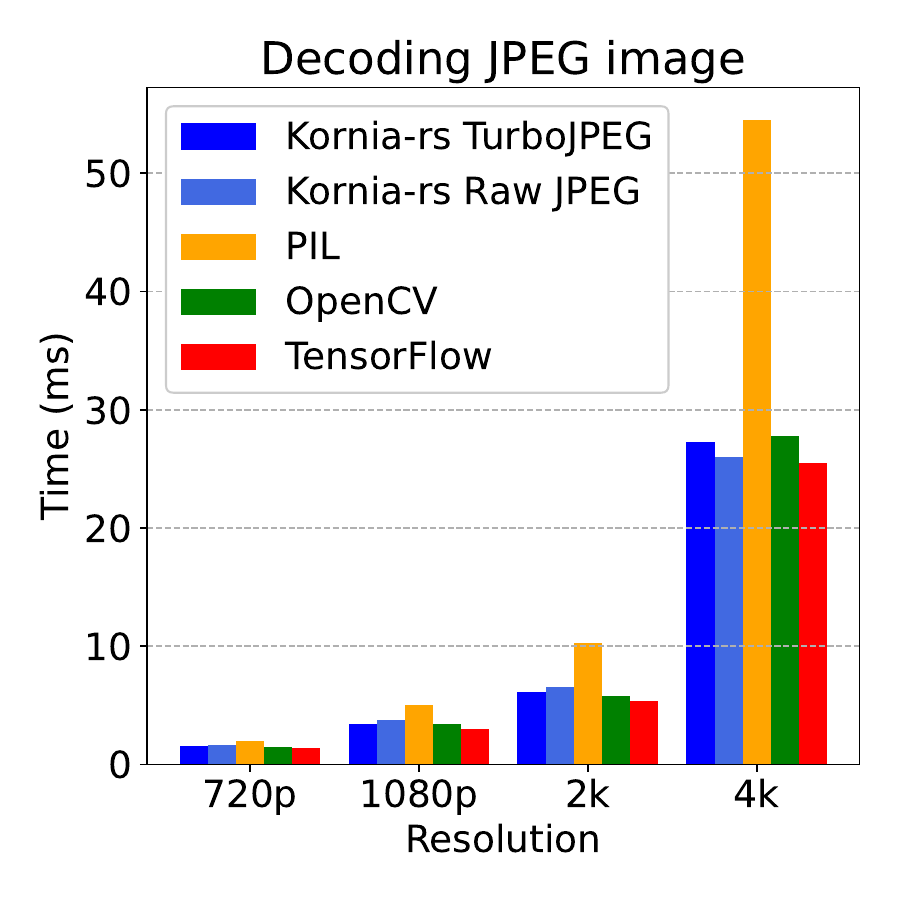}
        \hspace{.025\textwidth}
        \includegraphics[width=.3\textwidth]{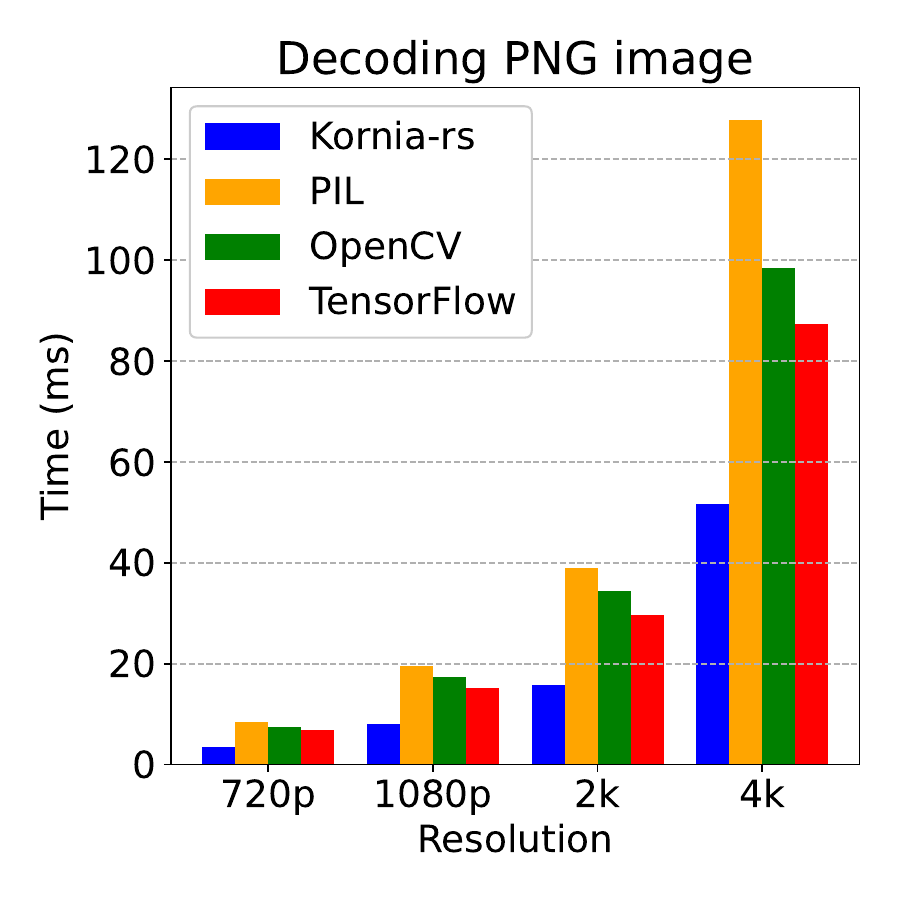}
        \hspace{.025\textwidth}
        \includegraphics[width=.3\textwidth]{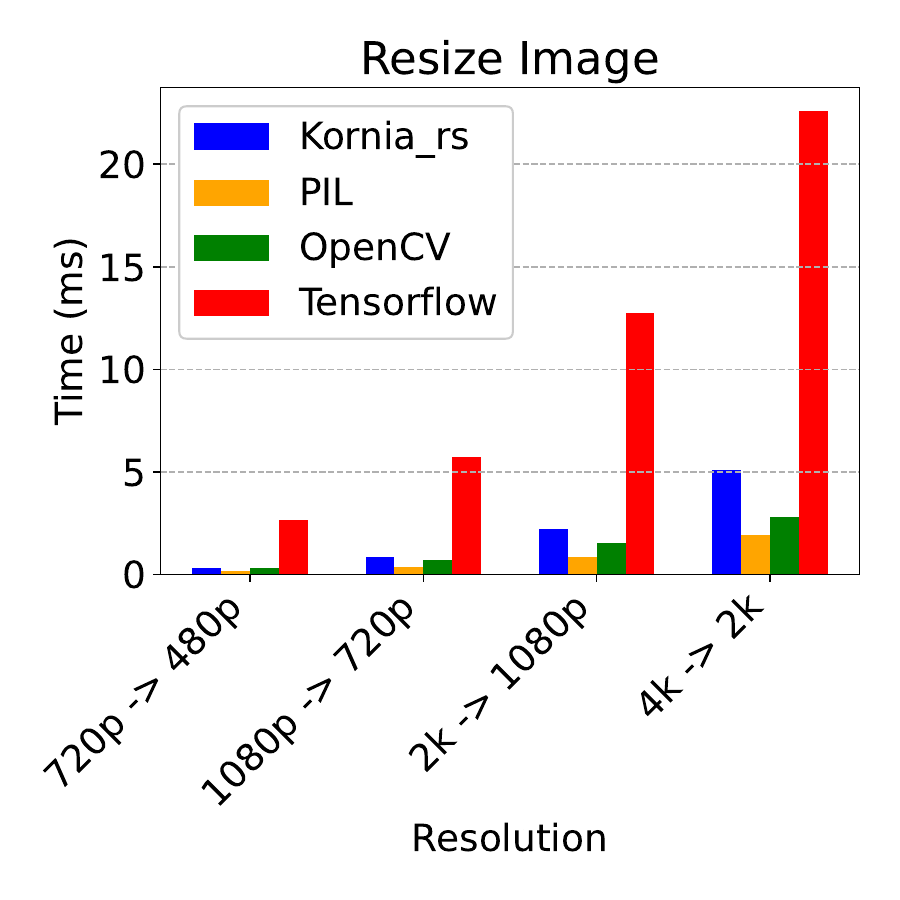}
        \caption{Comparison against C++ wrapper-based libraries using the Python bindings.}
        \label{fig:python-benchmarks}
    \end{subfigure}
    \caption{Performance demonstrations. Upper: We compare common image processing operations between \textit{kornia-rs} and \textit{image-rs}. Lower: We compare image I/O operations against OpenCV, PIL, and TensorFlow.}
\end{figure*}

\section{Performance Evaluation}

We compare \texttt{kornia-rs} against Rust~\cite{matsakis2014rust} libraries such as \texttt{image-rs} basic image processing operations, and popular CV libraries with Python bindings such as OpenCV~\cite{opencv_library}, PIL~\cite{Murray2025-dd}, and TensorFlow~\cite{tensorflow2015-whitepaper} for image I/O operations. Our results show that \textit{kornia-rs} achieves superior performance in both native Rust code and through its Python bindings. Each benchmark was run on an AMD Ryzen 7 5800X with 32GB RAM running Linux, Python 3.12, with results averaged over 1000 iterations.





\noindent\textbf{Image Processing Performance in Rust} As shown in \Cref{fig:rust-benchmarks}, \textit{kornia-rs} consistently outperforms \textit{image-rs}~\cite{image-rs} by a factor of $3\sim 5\times$ across all operations tested, including geometric transformations such as horizontal flip, oversizing, downsizing, as well as color space transformations such as RGB to grayscale.
The performance gap widens as image resolution increases, demonstrating \textit{kornia-rs}'s superior scaling behavior for high-resolution imagery.




\noindent\textbf{Python Bindings Performance} \Cref{fig:python-benchmarks} demonstrates that \textit{kornia-rs} delivers native Rust throughput through our python bindings with negligible overhead: it decodes PNG images end-to-end \textit{(raw bytes to \texttt{numpy.ndarray})} about $2.4\times$ faster than PIL and nearly $2\times$ faster than TensorFlow/OpenCV. Similarly, JPEG decoding is about $1.3-2\times$ faster than PIL and matches TensorFlow / OpenCV within $\pm 10\%$. The nearest-neighbor resizing is $4\sim 8\times$ faster than TensorFlow and matches the PIL/OpenCV throughput.

\begin{figure}[h]
\centering
\includegraphics[width=0.9\columnwidth]{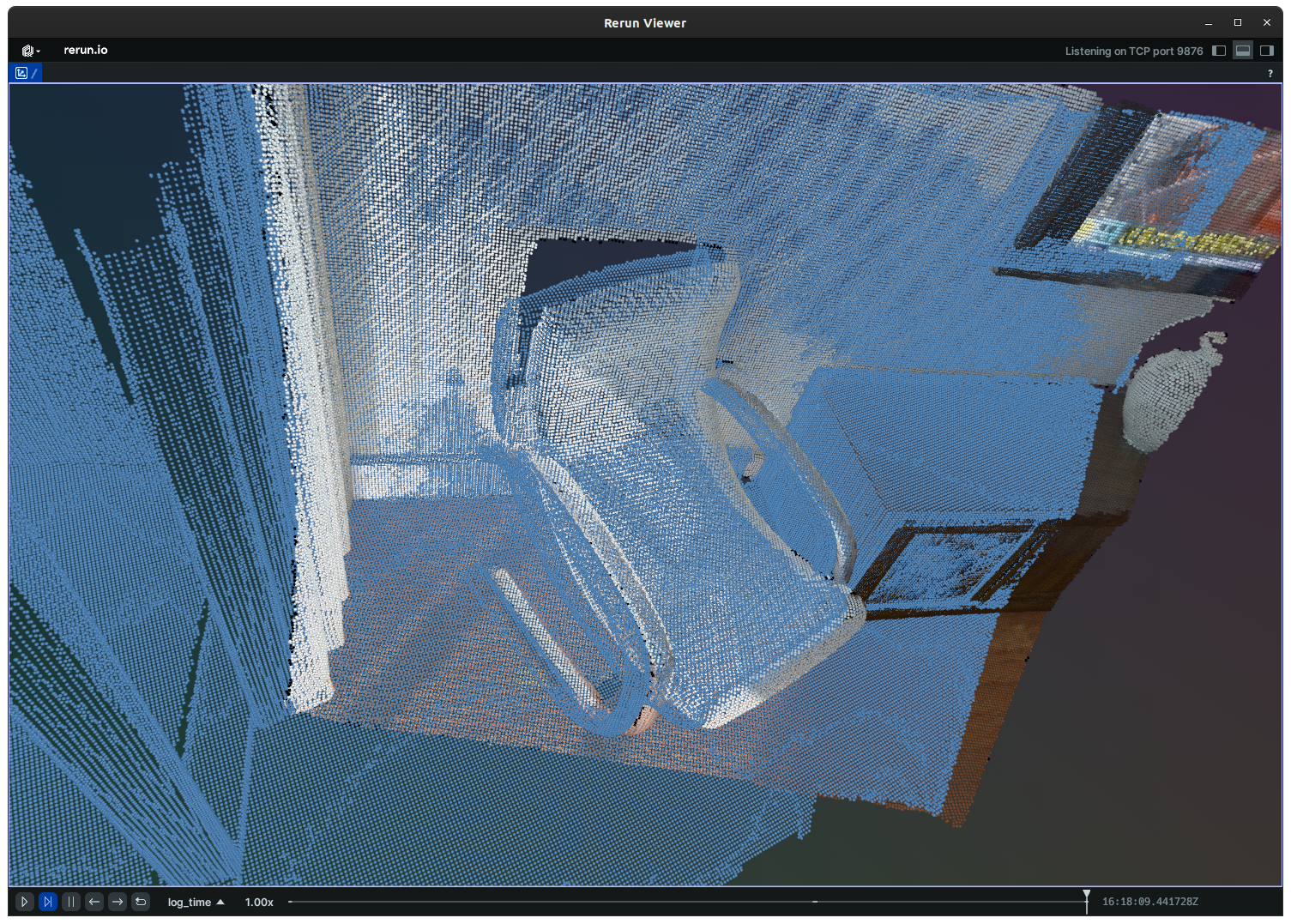}
\caption{Point cloud alignment using kornia-rs ICP implementation. The figure shows the source point cloud (rgb color) and the aligned result (blue).}
\label{fig:icp-alignment}
\end{figure}

\vspace{-10pt}

\section{Applications and Integrations}

This section demonstrates practical applications of \textit{kornia-rs} in real-world scenarios, highlighting its versatility and integration capabilities.

\subsection{Point Cloud Alignment via ICP}

\textit{Kornia-rs} provides a native implementation of the Iterative Closest Point (ICP) algorithm for 3D point cloud alignment. The module is built atop the library’s optimized tensor core, enabling high-throughput computations with support for pre-allocated buffers to eliminate runtime memory overhead. The implementation supports configurable error metrics and correspondence strategies, making it suitable for a variety of robotics and vision applications. It integrates seamlessly with the \texttt{kornia-3d} module for transformation utilities and is designed for deterministic behavior to ensure reproducibility.

\subsection{Camera stream application}

\textit{Kornia-rs} provides a high-level abstraction over the \textit{GStreamer} multimedia framework, enabling seamless access to a wide range of camera sources—from standard USB webcams to advanced GMSL industrial cameras and remote RTSP streams. This abstraction simplifies the integration of video input into computer vision pipelines, allowing developers to quickly prototype and deploy applications. The following example demonstrates how to acquire a stream from an RTSP camera, perform a basic color space conversion, and save the processed frames to disk.

{\footnotesize
\begin{lstlisting}[language=rust, basicstyle=\footnotesize\ttfamily,]
use kornia::{image::Image, io::functional as F};

fn main() -> Result<(), Box<dyn std::error::Error>> {
    // Create RTSP camera stream
    let mut capture = RTSPCameraConfig::new()
        .with_settings(
            &username,
            &password,
            &camera_ip,
            &camera_port,
            &stream_path
        )
        .build()?;
    // Start the camera stream
    capture.start()?;
    // Read and write frames
    while let Some(rgb) = capture.grab()? {
        // Preallocate output buffer
        let mut gray = Image::<u8, 1>::from_size_val(rgb.size(), 0)?;
        // execute the operator
        imgproc::color::gray_from_rgb(rgb, &mut gray)?;
        // Write result in disk
        F::write_image_jpeg("output.jpg", &gray)?;
    }
    Ok(())
}
\end{lstlisting}
}
\subsection{Integration with robotics frameworks}

\textit{kornia-rs} can integrate easily with Rust native robotics such as copper-rs~\cite{copper-rs} for building real-time systems. The following example demonstrates implementing a Sobel edge detection task:

{\footnotesize
\begin{lstlisting}[language=rust, basicstyle=\footnotesize\ttfamily,]
use cu29::prelude::*;
use kornia::{image::Image, imgproc};

impl<'cl> CuTask<'cl> for Sobel {
    type Input = input_msg!('cl, ImageRgb8Msg);
    type Output = output_msg!('cl, ImageGray8Msg);

    fn process(
        &mut self,
        _clock: &RobotClock,
        input: Self::Input,
        output: Self::Output,
    ) -> Result<(), CuError> {
        // Get input image from message
        let Some(img) = input.payload() else {
            return Ok(());
        };
        // Pre-allocate output buffer
        let mut img_sobel = Image::from_size_val(
            img.size(), 0
        )?;
        // Apply Sobel filter using kornia-rs
        imgproc::filter::sobel(&img, &mut img_sobel, 3)?;
        // Set the output payload
        output.set_payload(ImageGray8Msg(img_sobel));
        Ok(())
    }
}
\end{lstlisting}
}

\textit{Kornia-rs} can be embedded into larger systems while maintaining its performance and safety guarantees. In our GitHub repository\footnote{\url{https://github.com/kornia/kornia-rs.git}}, we further provide example integrations, \textit{e.g.} with dora-rs~\cite{dora-rs} which enables building complex robotics applications with real-time image processing capabilities.

\section{Future Work}

Future development of \textit{kornia-rs} will focus on three key areas: (1) expanded hardware acceleration through GPU-accelerated operators via CubeCL, enabling significant performance improvements for computational intensive tasks; (2) integration with deep learning models, particularly visual language models and visual language action models for robotics and embodied AI applications; and (3) implementation of additional classical computer vision algorithms such as SLAM and visual odometry. These directions aim to establish \textit{kornia-rs} as a comprehensive, high-performance solution for computer vision applications across diverse deployment environments.

\bibliographystyle{ieeetr}
\bibliography{reference}

\end{document}